\documentclass[authorversion, nonacm]{jarxiv}

\usepackage{graphicx}
\usepackage{amsmath}
\usepackage{enumitem}
\usepackage{times}

\usepackage{mathtools}
\usepackage{multirow}
\usepackage{adjustbox}
\usepackage{makecell}
\usepackage{xfrac}
\usepackage[algo2e, ruled]{algorithm2e}
\usepackage[textsize=footnotesize]{todonotes}
\usepackage{tikz}
\usepackage{makecell}
\usepackage{url}

\usetikzlibrary{automata, arrows.meta}

\definecolor{nodeborder}{RGB}{0,55,75}
\definecolor{greenedge}{RGB}{50,180,35}
\definecolor{blueedge}{RGB}{0,100,130}

\newtheorem{problem}{Problem}
\makeatletter
\renewcommand{\@algocf@capt@plain}{above}%
\makeatother

\acmVolume{0}
\acmArticle{0}
\acmMonth{0}
\acmYear{0}

\RequirePackage[
  datamodel=acmdatamodel,
  style=acmauthoryear,
  backend=biber,
  giveninits=true,
  uniquename=init
  ]{biblatex}

\begin{document}

\title[Active O2O-RL]{Active Offline-to-Online Reinforcement Learning}

\author{Alper Kamil Bozkurt}
\authornote{Corresponding Author.}
\email{bozkurta@vcu.edu}
\affiliation{%
  \institution{Virginia Commonwealth University}
  \city{Richmond}
  \state{Virginia}
  \country{USA}
}

\author{Shangtong Zhang}
\email{xdm2bt@virginia.edu}
\affiliation{%
  \institution{University of Virginia}
  \city{Charlottesville}
  \state{Virginia}
  \country{USA}
}

\author{Yuichi Motai}
\email{ymotai@vcu.edu}
\affiliation{%
  \institution{Virginia Commonwealth University}
  \city{Richmond}
  \state{Virginia}
  \country{USA}
}

\renewcommand{\shortauthors}{Bozkurt, Zhang \& Motai}

\begin{abstract}
\textbf{Background:} Offline reinforcement learning (RL) enables effective policies to be trained from large, previously collected datasets and subsequently improved through limited online interaction. This offline-to-online RL (O2O-RL) paradigm is particularly promising in nonstationary domains where interaction is costly or potentially hazardous. Standard O2O-RL pipelines train multiple candidate policies offline, evaluate them using off-policy or online evaluation, and then deploy and fine-tune the policy with the highest estimated value. However, as in offline pretraining, fine-tuning performance is highly sensitive to the choice of algorithm and hyperparameters, making it risky to commit to a single policy. \\
\textbf{Objectives:} We study active policy selection for fine-tuning under a limited interaction budget in O2O-RL settings. To our knowledge, this is the first work to address this problem. \\
\textbf{Methods:} We formulate the problem by identifying a fundamental trade-off between allocating online interactions to policy evaluation, which helps identify high-performing policies, and allocating them to fine-tuning, which improves policy performance. We then propose an approach that balances this trade-off by actively selecting policies for fine-tuning based on upper-confidence bounds on their future performance. These bounds are derived from locally linear performance forecasts fitted to observations obtained through online evaluation. \\
\textbf{Results:} Across a diverse range of experiments, the proposed approach consistently outperforms existing O2O-RL baselines. \\
\textbf{Conclusions:} Actively selecting and fine-tuning policies uses limited online interaction budgets more effectively than either committing to a single policy or dividing the budget equally among all policies. Our framework also advances offline RL toward practical deployment in real-world systems where online interaction is costly or risky.
\end{abstract}

\maketitle
\section{Introduction}
Reinforcement learning (RL) is becoming a key ingredient in the autonomy of modern robotic systems operating in unstructured, dynamic environments \cite{singh2022reinforcement}. By learning to make decisions and derive control actions directly from onboard sensing and perception, RL can substantially reduce human workload and the likelihood of human error \cite{zhang2022reinforcement}, thereby facilitating widespread real-world deployment. Deep RL has proven effective in synthesizing control policies for high-dimensional, nonlinear physical systems for which manual controller design is infeasible, leading to numerous successful applications \cite{tang2025deep}. Despite the flexibility and power of this framework, standard RL methods typically require extensive direct interaction with the physical environment for exploration \cite{ladosz2022exploration}. They are therefore impractical for training policies from scratch when such interactions are costly, risky, or time-consuming \cite{dulac2021challenges}.

Offline RL \cite{levine2020offline, prudencio2023survey} has emerged as an alternative to online RL, enabling policies to be trained from large, previously collected datasets. These datasets are usually collected under safe, controlled conditions \cite{zhou2023real}, often, though not exclusively, by human operators. A central challenge in offline RL is that the performance of a pretrained policy becomes unpredictable as its behavior diverges from that of the policy used to collect the dataset \cite{kostrikov2021offline}. Due to this distributional shift, a policy pretrained via offline RL can perform arbitrarily poorly in the real environment \cite{qin2022neorl}. To partially mitigate this issue, offline policy selection, usually performed via off-policy evaluation (OPE) \cite{paine2020hyperparameter, uehara2025review}, is used to identify high-performing policies among candidates pretrained using different hyperparameter configurations. However, OPE estimates are generally not sufficiently reliable to determine which policy should be deployed, primarily because they are also vulnerable to distributional shift \cite{brandfonbrener2021offline}. This difficulty is further exacerbated by nonstationary real-world environments that require online adaptation \cite{julian2021never}.

The need to evaluate and fine-tune policies pretrained via offline RL through online interaction has motivated the offline-to-online RL (O2O-RL) paradigm, which treats the entire pipeline as a single joint problem. By combining offline pretraining with a small amount of online interaction, O2O-RL can efficiently produce high-performing policies, thereby enabling the broader deployment of RL in physical domains. As a result, O2O-RL has attracted substantial attention and led to the development of numerous methods. Most O2O-RL approaches (e.g., \citet{nair2020awac, lee2022offline, ball2023efficient}) focus on improving the offline stage to produce pretrained policies that adapt more effectively to real environments during fine-tuning. Although these methods improve overall performance, they do not address the fundamental sensitivity of policy performance to hyperparameter choices and environmental conditions. More recently, several methods have examined policy selection in O2O-RL (e.g., \citet{konyushova2021active, kurenkov2022showing}), aiming to identify the best candidate by evaluating pretrained policies under a limited online interaction budget. However, these methods do not incorporate fine-tuning into the policy-selection process and overlook the trade-off arising from allocating the same limited interaction budget to both policy evaluation and fine-tuning.

\begin{figure}
    \centering
    \includegraphics[width=0.8\linewidth]{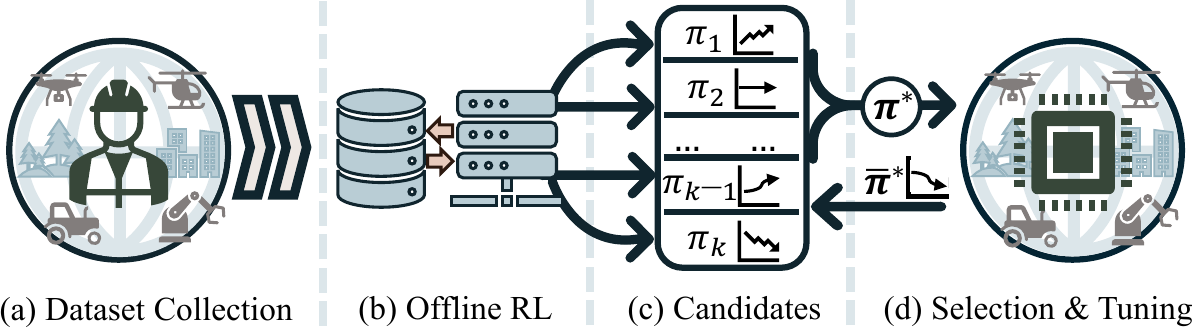}
    \caption{\textbf{Proposed O2O-RL Framework.} (a) Datasets are typically collected in controlled but imperfect settings. (b) Offline RL is used to train a diverse set of candidate policies using different algorithms and hyperparameter configurations. (c) A local linear model predicts the future performance of each policy and constructs an upper confidence bound (UCB). (d) The policy with the highest UCB is selected and fine-tuned, after which its predicted future performance and UCB are updated. Whenever its UCB falls below that of another policy, it is replaced by the policy with the higher UCB.}
    \label{fig:framework}
\end{figure}

In this work, we address the problem of obtaining high-performing, deployable policies in O2O-RL from a broader and more practical perspective. Prior work and our experiments show that pretrained policies can perform arbitrarily poorly in real environments and that fine-tuning may require substantial interaction to improve performance. Fine-tuning can even cause performance regressions, with outcomes varying across algorithms, hyperparameters, and environments. To manage this performance volatility under a limited interaction budget, we propose an O2O-RL approach (Fig.~\ref{fig:framework}) that jointly performs active policy selection and fine-tuning. During the offline stage, we first train a diverse set of candidate policies using offline RL across a representative range of algorithms and hyperparameter configurations. During the novel online stage, we actively select and fine-tune candidate policies using an upper-confidence-bound (UCB) criterion based on their predicted future performance until the interaction budget is exhausted. After each online episode, we fine-tune the selected policy, evaluate its performance, and then fit a local linear model to forecast its future performance and construct a corresponding UCB. We switch policies whenever the bound of the selected policy falls below that of another candidate. In this manner, we efficiently allocate the interaction budget to both identifying promising policies and improving them through fine-tuning.

We summarize our main contributions as follows:
\begin{itemize}
\item To the best of our knowledge, we are the first to study policy selection for fine-tuning under a limited interaction budget in O2O-RL settings.
\item We propose a new approach that actively selects and fine-tunes policies using a UCB criterion based on future performance predicted through local regression.
\item We evaluate our approach on a suite of simulated robotics benchmarks and demonstrate consistent gains over standard O2O-RL baselines.
\end{itemize}

This article substantially extends our earlier conference paper \cite{bozkurt2026adaptiveO2ORL} by providing additional methodological and technical discussion, a more comprehensive analysis of the proposed framework, and considerably expanded experiments. In particular, this version includes results on a broader collection of navigation, classic-control, and locomotion tasks, as well as ablation studies evaluating the method's sensitivity to its design choices and its generality across environments and experimental settings.

\section{Related Work}

\subsection{Offline Reinforcement Learning} 

\subsubsection{Basic Approaches}
A naive approach to offline RL is to train policies via standard online, off-policy algorithms such as deep deterministic policy gradient (DDPG) \cite{lillicrap2016continuous}, twin delayed DDPG (TD3) \cite{fujimoto2018addressing}, and soft actor-critic (SAC) \cite{haarnoja2018soft} on batches drawn from previously collected datasets. However, such policies can fail catastrophically, as they may lead to states outside the dataset due to substantially inaccurate value estimates arising from distributional shift.
In contrast, behavior cloning (BC); e.g., \cite{torabi2018behavioral, florence2022implicit}, which aims to imitate the dataset behavior and thereby largely prevent distributional shift, often cannot surpass the performance of the behavior policy, which is undesirable for non-expert datasets.

\subsubsection{Offline RL for Distributional Shift}
The central objective in offline RL is to improve upon the behavior policy of the dataset while avoiding severe failures caused by distributional shift. Most offline RL methods therefore balance this trade-off by permitting off-policy updates with additional regularization or constraints. Some approaches, such as batch-constrained deep Q-learning (BCQ) \cite{fujimoto2019off}, bootstrapping error accumulation reduction (BEAR) \cite{kumar2019stabilizing}, and policy in the latent action space (PLAS) \cite{zhou2021plas}, focus on constraining actions to prevent policies from taking actions outside the dataset. Other approaches, such as conservative Q-learning (CQL) \cite{kumar2020conservative} and TD3+BC \cite{fujimoto2021minimalist}, incorporate, respectively, Q-value and BC regularization terms into the policy updates. We refer readers to \cite{levine2020offline} and \cite{prudencio2023survey} for comprehensive reviews of offline RL approaches.

\subsection{Offline-to-Online Reinforcement Learning}

\subsubsection{Offline RL for Fine-Tuning}
The fundamental limitation that the performance of offline RL methods is upper-bounded by dataset quality necessitates fine-tuning pretrained policies via online interactions. This insight has motivated methods that treat the entire offline-to-online framework as a unified problem, where the main objective is to improve the final performance of policies after fine-tuning. Advantage weighted actor critic (AWAC) \cite{nair2020awac} improves the efficiency of fine-tuning with off-policy data via dynamic programming. Implicit Q-learning (IQL) \cite{kostrikov2021offline} utilizes a policy-extraction step during training that aids fine-tuning. Calibrated Q-learning (CalQL) \cite{nakamoto2023cal}, building on CQL, learns conservative Q-values better suited for fine-tuning. Revisited behavior regularized actor-critic (ReBRAC) \cite{tarasov2023revisiting} extends TD3+BC with simple hyperparameter choices that benefit both offline training and fine-tuning. Hybrid RL (Hy-Q) \cite{song2023hybrid} augments offline and online data and applies a fitted Q-iteration procedure. Robust O2O \cite{wen2024towards} uses an ensemble of Q-networks and smoothness regularization to mitigate distributional shift. In this work, we do not attempt to introduce or improve any particular O2O-RL algorithm. Instead, we treat the choice of algorithm as a high-level hyperparameter when training candidate policies and fine-tuning, since no single O2O-RL method is uniformly best in all environments.

\subsubsection{Online Policy Selection}
Recent studies, \cite{konyushova2021active} and \cite{kurenkov2022showing}, studied the policy selection problem in O2O-RL. In a setting similar to ours, these studies use a limited online-interaction budget to identify the best-performing pretrained policies, rather than relying on OPE estimates as in prior offline RL approaches \cite{paine2020hyperparameter, uehara2025review}. Since evaluating all candidates online would quickly exhaust the budget, they propose active selection strategies to decide which policies to evaluate, thereby allocating the budget efficiently. In contrast, our focus is to allocate online interactions to identify the policies that will perform best after fine-tuning under the given budget.

\section{Problem Formulation}

\subsection{Background and Setup}
Following the standard RL framework, we model interactions with an environment to perform tasks as Markov decision processes (MDPs). Formally, an MDP $\mathcal{M}$ is a tuple $(S,A,P,p_0,R,\gamma)$, where $S$ is the state set, $A$ is the action set, $P$ and $p_0$ are the transition and initial-state distributions, $R$ is the reward function, and $\gamma$ is the discount factor. The ultimate objective in RL is to obtain a policy $\pi:S\mapsto A$ that maximizes the expected cumulative discounted reward; i.e., 
\begin{align}
    \pi^* \coloneqq \mathrm{argmax}_\pi \mathbb{E}\left[\sum_{t=0}^\infty \gamma^t r_t\right].
\end{align}
In offline RL, this objective must be achieved using a dataset $\mathcal{D}$ of transitions $\{(s_i,a_i,r_i,s'_i)\}_{i=1}^D$ collected by a typically unknown behavior policy $\pi_b$ in $\mathcal{M}$, without further interaction with the environment.

In this work, we extend the offline RL setting by allowing $M$ online interactions in addition to the dataset $\mathcal{D}$; i.e., a set of at most $M$ new transitions may be collected from the environment. Unlike in \cite{konyushova2021active} and \cite{kurenkov2022showing}, in our O2O-RL setting, these online interactions can be used not only for evaluation but also for fine-tuning, introducing a novel trade-off. For simplicity, we represent the budget as $N$ fine-tuning \& evaluation iterations, each using $M/N$ transitions. Our aim is then, given an MDP $\mathcal{M}$, a dataset $\mathcal{D}$ collected from $\mathcal{M}$, and an interaction budget $N$, to obtain the highest-performing policy achievable, in the sense of minimizing regret.

\subsection{Problem Statement}
We formalize policy selection as the following simple regret-minimization problem:

\begin{problem} \label{problem}
Consider a set of $K$ candidate policies pretrained offline on a dataset with different hyperparameter settings, and let $\pi^i_j$ denote the policy that is obtained from the $i$th pretrained policy after $j$ fine-tuning   iterations, and let 
\begin{align}
    v^i_j \coloneqq \mathbb{E}_{\pi^i_j}\left[\sum_{t=0}^\infty \gamma^t r_t\right]
\end{align}
denote its corresponding value. For a given interaction budget $N$, the optimal selection is then a policy $\pi^{i^*}_{j^*}$ such that 
\begin{align}
    i^*, j^* \coloneqq \underset{1\leq i\leq K,0\leq j\leq N}{\mathrm{argmax}} v^i_j.
\end{align}
Our objective is to devise a procedure that fine-tunes each pretrained policy $\pi^i_0$ for $\bar{j}_i$ iterations and estimates the values $\{\hat{v}^i_0, \dots, \hat{v}^i_{\bar{j}_i}\}$ under the constraint $\sum_{i=1}^K \bar{j}_i \leq N$ such that the value of the selected policy 
\begin{align}
    \hat{v}^* = \max\limits_{1\leq i \leq K} \max\limits_{0\leq j\leq \bar{j}_i} \hat{v}^i_j    
\end{align}
minimizes
\begin{align}
    \textrm{regret} \coloneqq v^{i^*}_{j^*} - \hat{v}^*.
\end{align}
\end{problem}

In our setting, each policy $\pi^i_j$ is specified by trained parameters $\theta^i_j$, an RL algorithm $\mathcal{A}^i$, and a set of hyperparameters $\lambda^i$ that determine the offline training, fine-tuning, and action-selection procedures. Each configuration $\psi^i =(\mathcal{A}^i,\lambda^i)$, indexed by $i \in \{1,\dots,K\}$, is predetermined and fixed, and is used to produce a policy lineage $\pi^i_0, \pi^i_1, \dots$ with corresponding parameters $\theta^i_0, \theta^i_1, \dots$. Here, $\theta^i_0$ is obtained through offline training, and for $j \geq 1$, $\theta^i_j$ is obtained from $\theta^i_{j-1}$ after one fine-tuning iteration. Henceforth, we omit the explicit notation for $\mathcal{A}^i$, $\lambda^i$, and $\theta^i_j$, and instead treat them as implicit in the indices of $\pi^i_j$.
The optimal selection $\pi^{i^*}_{j^*}$ intuitively represents the best policy that could have been obtained if the policy sequences and their corresponding values were known a priori. It therefore constitutes an upper bound on the improvement achievable by any policy-selection procedure during fine-tuning.

We can establish a lower bound on the regret of this formulation in a general setting as a function of the number of candidates $K$ and the budget $N$. Suppose that the estimates of the policy values follow a normal distribution with mean $\mu$ and standard deviation $\sigma$ for every configuration index $i$ and fine-tuning iteration $j$, independently of all previous values. The expected maximum value $\mathbb{E}[v^{i^*}_{j^*}]$ that can be obtained by fine-tuning all of these policies for $N$ iterations can be lower-bounded as follows:
\begin{align}
    \mathbb{E}[v^{i^*}_{j^*}] &= \mathbb{E}\left[\max\limits_{1\leq i\leq K,0\leq j\leq N} \hat{v}^i_j\right] \\
    &\geq  \sigma \sqrt{\frac{\log(K(N+1))}{\pi\log(2)}}.
\end{align}
In this setting, all selection procedures are equivalent to choosing the first configuration for fine-tuning at every iteration, and they yield the same expected value, which is upper-bounded by
\begin{align}
    \mathbb{E}[\hat{v}^*] &= \mathbb{E}\left[\max\limits_{0\leq j\leq N} v^1_j\right] \\
    &\leq  \sigma \sqrt{2\log(N+1)}.
\end{align}
Thus, we obtain the following lower bound on the regret:
\begin{align}
    \mathbb{E}[\textrm{regret}] &= \mathbb{E}[v^{i^*}_{j^*} - \hat{v}^*] \\
    &= \mathbb{E}[v^{i^*}_{j^*}] - \mathbb{E}[\hat{v}^*] \\
    & \geq \sigma \left(\sqrt{\frac{\log(K(N+1))}{\pi\log(2)}} - \sqrt{2\log(N+1)} \right)
\end{align}
This lower bound becomes positive for $K=(N+1)^q$ with $q> 2\pi\log(2) - 1 \approx 1.95$ and increases as $K$ increases. This issue stems from the fact that the number of pretrained policies $K$ is given rather than chosen in the problem formulation, which causes the regret to increase inherently with $K$. However, we implicitly assume that $K \leq N$, which alleviates this limitation.

\subsection{Relation to Bandit and Low-Switching-Cost Formulations}

Our formulation is closely related to pure-exploration multi-armed bandits \cite{kaufmann2016complexity}, where the objective is to identify the best arm under a fixed budget without considering cumulative regret. In our formulation, arms correspond to configurations, playing an arm corresponds to selecting the most recent fine-tuned policy from the corresponding configuration, and the payoff is the value of the resulting policy after fine-tuning. The key distinction is that, in our setting, the arms are nonstationary, since the policy lineages induced by configurations evolve over time through fine-tuning. Moreover, the objective is to obtain the highest-payoff play, namely, to identify the fine-tuned policy with the highest value.

Our formulation is also related to the notions of low switching cost \cite{bai2019provably} and deployment complexity \cite{huang2022towards}, where the objective is to minimize the number of policy updates required to obtain a near-optimal policy. However, these formulations differ from ours in a key respect: we do not directly address the design of a new RL algorithm, nor do we aim to find a near-optimal policy with respect to cumulative reward. Instead, we focus on selecting, from a given set of candidate policies, the policy that achieves the highest performance after fine-tuning under a limited interaction budget.

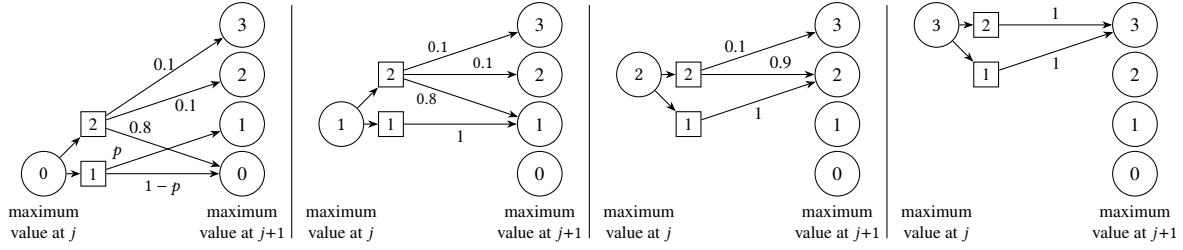
\begin{figure*}
\centering
\resizebox{0.99\textwidth}{!}{
    \begin{tikzpicture}[>=Stealth,->,semithick,
        every state/.style={draw,circle,minimum size=9mm,inner sep=0pt, font=\large},
        small/.style={draw,rectangle,minimum size=5mm,inner sep=0pt}
    ]
        \node[state] (v0) at (0,0) {$0$};
        \node[small] (i10) at (1,0) {1};
        \node[small] (i20) at (1,1) {2};
        
        \node[state] (v00) at (4,0) {0};
        \node[state] (v10) at (4,1) {1};
        \node[state] (v20) at (4,2) {2};
        \node[state] (v30) at (4,3) {3};

        \node (j00) at (0, -1) {\makecell{maximum\\value at $j$}};
        \node (j00) at (4, -1) {\makecell{maximum\\value at $j{+}1$}};

        \path (v0) edge node {} (i10);
        \path (v0) edge node {} (i20);

        \path (i10) edge node[below] {$1-p$} (v00);
        \path (i10) edge node[pos=0.1, above] {$p$} (v10);

        \path (i20) edge node[pos=0.3, above] {0.8} (v00);
        \path (i20) edge node[pos=0.7, below] {0.1} (v20);
        \path (i20) edge node[above] {0.1} (v30);

        \draw[-] (5,-1.3) -- (5,3.4);

        \node[state] (v1) at (6,1) {$1$};
        \node[small] (i11) at (7,1) {1};
        \node[small] (i21) at (7,2) {2};
        
        \node[state] (v01) at (10,0) {0};
        \node[state] (v11) at (10,1) {1};
        \node[state] (v21) at (10,2) {2};
        \node[state] (v31) at (10,3) {3};

        \node (j00) at (6, -1) {\makecell{maximum\\value at $j$}};
        \node (j00) at (10, -1) {\makecell{maximum\\value at $j{+}1$}};

        \path (v1) edge node {} (i11);
        \path (v1) edge node {} (i21);

        \path (i11) edge node[below] {1} (v11);

        \path (i21) edge node[pos=0.2, below] {$0.8$} (v11);
        \path (i21) edge node[pos=0.7, above] {$0.1$} (v21);
        \path (i21) edge node[pos=0.3, above] {$0.1$} (v31);

        \draw[-] (11,-1.3) -- (11,3.4);

        \node[state] (v2) at (12,2) {$2$};
        \node[small] (i12) at (13,1) {1};
        \node[small] (i22) at (13,2) {2};
        
        \node[state] (v02) at (16,0) {0};
        \node[state] (v12) at (16,1) {1};
        \node[state] (v22) at (16,2) {2};
        \node[state] (v32) at (16,3) {3};

        \node (j00) at (12, -1) {\makecell{maximum\\value at $j$}};
        \node (j00) at (16, -1) {\makecell{maximum\\value at $j{+}1$}};
        
        \path (v2) edge node {} (i12);
        \path (v2) edge node {} (i22);

        \path (i12) edge node[below] {1} (v22);

        \path (i22) edge node[pos=0.7, above] {0.9} (v22);
        \path (i22) edge node[pos=0.3, above] {0.1} (v32);

        \draw[-] (17,-1.3) -- (17,3.4);

        \node[state] (v3) at (18,3) {$3$};
        \node[small] (i13) at (19,2) {1};
        \node[small] (i23) at (19,3) {2};
        
        \node[state] (v03) at (22,0) {0};
        \node[state] (v13) at (22,1) {1};
        \node[state] (v23) at (22,2) {2};
        \node[state] (v33) at (22,3) {3};

        \node (j00) at (18, -1) {\makecell{maximum\\value at $j$}};
        \node (j00) at (22, -1) {\makecell{maximum\\value at $j{+}1$}};
        
        \path (v3) edge node {} (i13);
        \path (v3) edge node {} (i23);

        \path (i13) edge node[below] {1} (v33);

        \path (i23) edge node[ above] {1} (v33);
    \end{tikzpicture}
}
\caption{Illustration of a simple policy-selection problem for fine-tuning, formulated as a Markov decision process. The circles denote states representing the maximum policy value achieved at the current fine-tuning iteration; the rectangles denote choices among training configurations; and the numbers above the arrows indicate the corresponding transition probabilities.}
\label{fig:problem}
\end{figure*}

\subsection{Illustrative Example}
We illustrate two key factors induced by this problem formulation that affect the policy selection process: (i) the size of the remaining budget and (ii) the highest value among previously obtained fine-tuned policies.
Consider the example shown in Fig.~\ref{fig:problem}, where there are two candidate policies with configurations indexed by $i \in {1,2}$. Each fine-tuning iteration $j$, independently of previous fine-tuning iterations, produces a policy whose value $v$ is sampled from a static distribution associated with the corresponding training configuration. For simplicity, we assume that these distributions and their parameters are known a priori, and that policy values can be obtained directly without online evaluation.

Fine-tuning the policy with the second configuration ($i=2$) exhibits higher variance than fine-tuning the policy with the first configuration ($i=2$). Specifically, for $i=1$, the policy value is $1$ with probability (w.p.) $p$ and $0$ w.p. $1-p$; for $i=2$, the value is $3$ w.p. $0.1$, $2$ w.p. $0.1$, and $0$ w.p. $0.8$. The expected values are therefore $p$ and $0.5$, respectively.
Now consider the case $N=3$. The optimal strategy at any iteration $j$ can be specified as a function of $v_j^*$, the value of the best-performing fine-tuned policy obtained up to iteration $j$. If $v_j^*=3$, selecting either configuration is optimal because the highest possible value has already been obtained. If $v_j^* \in \{1,2\}$, the optimal strategy is to choose $i=2$, since it is the only configuration that can produce a policy with a higher value.

When $v_j^*=0$, the optimal strategy depends strongly on both $j$ and $p$. For $p<0.5$, the optimal strategy is to choose $i=2$ at every iteration, yielding $\alpha^*=\{2,2,2\}$ for $j\in\{0,1,2\}$, because its expected value is higher than that of $i=1$ regardless of $v_j$. When $p=0.6$, the optimal strategy becomes $\alpha^*=\{1,1,1\}$ as the expected value of $i=1$ increases. However, further increasing $p$ to $0.9$ and $0.99$ yields the optimal strategies $\alpha^*=\{2,1,1\}$ and $\alpha^*=\{2,2,1\}$, respectively. This occurs because larger values of $p$ make it increasingly likely that a fine-tuned policy with value $1$ can be obtained in later iterations, which encourages greater risk taking in earlier iterations in pursuit of policies with values $2$ and $3$. This example illustrates the distinctive nature of our problem formulation.

\section{Active Offline-to-Online Reinforcement Learning}

\begin{figure}
    \centering
    \includegraphics[width=\linewidth]{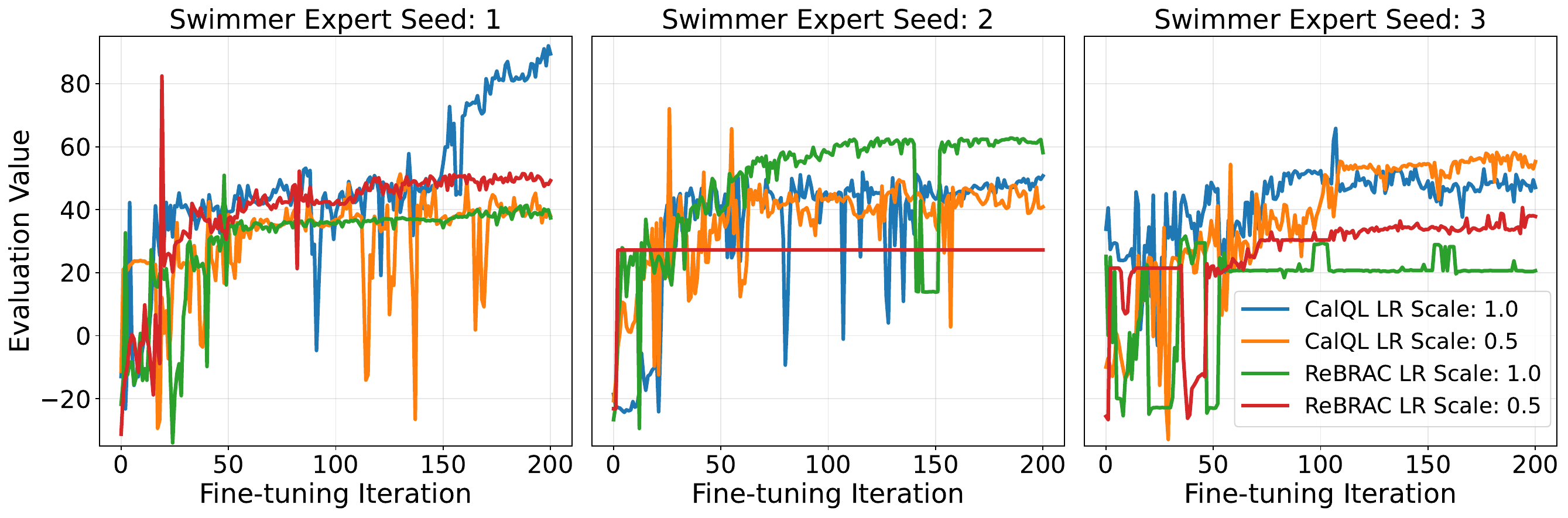}
    \caption{Evolution of the return values of pretrained policies during fine-tuning on \textsc{swimmer-expert} for three random seeds, each of which determines the initial state. Policies are pretrained with offline RL algorithms CalQL and ReBRAC for 200 K steps using default and half-default learning rates (LR). The value curves are highly irregular: they may improve, stall, regress after initial improvement, or exhibit high variance. Even with the same algorithms and hyperparameters, changing the initial state can significantly alter the outcome of offline training and the progression of the value curves, which necessitates an active policy selection approach during fine-tuning.}
    \label{fig:finetuning}
\end{figure}

\LinesNumbered
\SetAlCapHSkip{0pt}
\setlength{\algomargin}{1em}

\begin{algorithm2e*}[h]
\caption{Active Offline-to-Online Reinforcement Learning}\label{alg:active}
\SetKwInOut{Input}{Input}\SetKwInOut{Output}{Output}
\SetKwComment{Comment}{}{}
\Input{MDP $\mathcal{M}$, dataset $\mathcal{D}$, interaction budget $N$, window size $w$}
\Output{policy $\pi^*$}
\small
\Comment{\footnotesize}
\Comment{\# Offline Stage}
Estimate behavior policy value $\hat{v}_B$ using $\mathcal{D}$\\
Train $K$ candidate policies $\{\pi_0^i\}_{i=1}^K$ on $\mathcal{D}$ \Comment{{\footnotesize \ \# using diverse algorithms and hyperparameters}}
Initialize a policy lineage list $\Pi^{i}{\leftarrow}(\pi^i_0)$ for each $\pi_0^i$\\

\Comment{}
\Comment{\# Online Initialization}
$\pi^* \leftarrow$ null \Comment{\footnotesize{\ \# best policy}}
$v^* \leftarrow -\infty$ \Comment{\footnotesize{\ \# best value estimate}}
\For{$i=1$ \KwTo $K$}{
    Initialize a value estimate list $\hat{V}^{i} = \{\hat{v}_R, \hat{v}_R\}$\\
    $\pi^*, \hat{v}^* \leftarrow$ finetune\_eval($\mathcal{M}, \Pi^i, \hat{V}^i, \pi^*, \hat{v}^*$)\\
    $\tilde{v}^i, \tilde{s}^i \leftarrow$ forecast($\hat{V}^i, w, d, c$)
}
\Comment{}
\Comment{\small{\# Policy Selection \& Fine-Tuning}}
\For{$j=\tau \cdot K$ \KwTo $N$}{
    $i^* \leftarrow \min_i (v^* - \tilde{v}^i)/\tilde{s}^i$  \Comment{\footnotesize{\ \# pick the one requiring min scale to reach $v^*$}}
    $\pi^*, \hat{v}^* \leftarrow$ finetune\_eval($\mathcal{M}, \Pi^{i^*}, \hat{V}^{i^*}, \pi^*, \hat{v}^*$)\\
    $\tilde{v}^{i^*}, \tilde{s}^{i^*}, u^{i^*} \leftarrow$ forecast($\hat{V}^{i^*}, w, \min(d, N-j), c$)
}
\Return{$\pi^*$}
\end{algorithm2e*}

\begin{algorithm2e*}[h]
\caption{Fine-Tuning \& Online Evaluation}\label{alg:ft_eval}
\SetKwInOut{Input}{Input}\SetKwInOut{Output}{Output}
\SetKwProg{Fn}{function}{ finetune\_eval($\mathcal{M}, \Pi, \hat{V}, \pi^*, \hat{v}^*$)}{end}
\SetKwComment{Comment}{}{}
\Input{MDP $\mathcal{M}$, policy lineage list $\Pi$, value estimate list $\hat{V}$, best policy $\pi^*$, best value estimate $v^*$}
\Output{new best policy $\pi^*$, new best value estimate $v^*$}
\small
\Fn{}{
    $\pi_{t+1} \leftarrow \mathrm{finetune}(\mathcal{M}, \pi_t)$\\
    $\Pi\mathrm{.append}(\pi_{t+1})$\\
    $\hat{v}_{t+1} \leftarrow \mathrm{evaluate}(\mathcal{M}, \pi_{t+1})$\\
    $\hat{V}\mathrm{.append}(\hat{v}_{t+1})$\\
    \If{$\hat{v}_{t+1}>v^*$}{
        $v^* \leftarrow \hat{v}_{t+1}$\\
        $\pi^* \leftarrow \pi_{t+1}$
    }
}

\Return{$\pi^*$, $v^*$}
\end{algorithm2e*}

\begin{algorithm2e*}[h]
\caption{Value Forecast}\label{alg:val_forecast}
\SetKwInOut{Input}{Input}\SetKwInOut{Output}{Output}
\SetKwProg{Fn}{function}{ forecast($\hat{V}, w, \delta, c$)}{end}
\SetKwComment{Comment}{}{}
\Input{value estimate list $\hat{V}$, window size $w$, future iterations $\delta$, exploration degree $c$}
\Output{forecasted value $\tilde{v}$, prediction interval $\tilde{s}$}
\small

\Fn{}{
    $t \leftarrow \hat{V}\mathrm{.length}()$ \Comment{\footnotesize{\ \# number of iterations used to fine-tune}}
    $\omega \leftarrow \min\{w, t\}$ \Comment{\footnotesize{\ \# clipped window size}}
    Fit a local linear model $\hat{b}_t, \hat{m}_t$ to $\hat{V}[-\omega:]$\Comment{\footnotesize{\ \# last $\omega$ value estimates}}
    $t' \leftarrow t + \delta$ \Comment{\footnotesize{\ \# predictor}}
    Forecast the future value $\tilde{v}^i \leftarrow \hat{b}_t + \hat{m}_t t'$ with the prediction interval $\tilde{s}$
}
\Return{$\tilde{v}$, $\tilde{s}$}

\end{algorithm2e*}

Our overall procedure is shown in Algorithm \ref{alg:active}. We now explain our approach in detail below.

\subsection{Value Forecast}

We follow the offline initialization step in the standard O2O pipeline. We begin by training a diverse pool of $K$ candidate policies, $\{\pi_0^1,\dots,\pi_0^K\}$, using offline RL on the given dataset $\mathcal{D}$. We sweep across multiple algorithmic families and hyperparameter ranges to cover the design space suitable for the environment $\mathcal{M}$. Each candidate policy $\pi_0^i$ is associated with a configuration $\psi^i$ that contains all information needed for pretraining, fine-tuning, and prediction. This includes the offline and online RL algorithm choices, as well as all hyperparameters, such as the neural network architecture, activation functions, optimizers, loss functions, number of pretraining steps, exploration rate, random seed, and other relevant details. A configuration $\psi^i$, together with pretrained parameters $\theta_j^i$ obtained according to $\psi^i$, completely defines the candidate policy $\pi_0^i$.

At initialization, we use $\tau$ pseudo-estimates $(\hat{v}^i_{-\tau},\dots,\hat{v}^i_{0})$ for each configuration, where $\tau$ is chosen according to the complexity of the local regression model used for forecasting. For example, since we use linear regression in our experiments, we set $\tau=1$. These pseudo-estimates mitigate the initial identification and high-variance problems that arise when only a few early value estimates are available for fitting the model. We set all pseudo-estimates equal to the value of a random policy, $\hat{v}_{R}$, which we assume to be known a priori. We do not use OPE values for initialization because their accuracy can be quite limited.

The values of policies during fine-tuning do not always increase monotonically. They may stall or regress, exhibit sudden drops or jumps, and pass through both high- and low-variance regimes as we show in Fig.~\ref{fig:finetuning}. We adopt a local linear regression model to capture short-term upward or downward trends, together with a local residual variance estimate to capture time-varying uncertainty. Although we favor this minimalist design to highlight the key aspects of our approach, richer forecasting models, such as higher-order or weighted local polynomial regression, can also be employed depending on the application.

In particular, for each configuration $i$, we assume that the evolution of the value estimates during fine-tuning can be locally approximated by a linear model around time $t$:
\begin{align}
    v_t^i(t') \approx b^i_t + m^i_t\cdot t' + \epsilon^i_t(t')
\end{align}
where $\epsilon_t^i(t') \sim N(0,\sigma_t^i)$ is an independent and identically distributed normal error term. After obtaining each value estimate $\hat{v}_t^i$ during fine-tuning, we estimate the model parameters as follows:
\begin{align}
    \hat{b}^i_t, \hat{m}^i_t \coloneqq \underset{b, m}{\mathrm{argmin}} \sum_{t'= t-w}^{t} (\hat{v}^i_{t'} - b - m t')
\end{align}
where $w$ is the bandwidth parameter controlling the size of the window over which linearity is assumed to hold. We then use these estimated parameters to predict the future value at time $t'$:
\begin{align}
    \tilde{v}_t^i(t') = \hat{b}^i_t + \hat{m}^i_t t'.
\end{align}
We also calculate the prediction interval $\tilde{s}_t^i(t')$ as
\begin{align}
    \tilde{s}^i_t(t') \coloneqq& s^i_t \sqrt{1 + \frac{1}{w+1} + \frac{(t'-\bar{w})^2}{ws_w^2}}
\end{align}
where $s^i_t \coloneqq \sum_{t'= t-w}^{t} (\hat{v}^i_{t'} - \hat{b}^i_t, - \hat{m}^i_t, t')/(w-1)$ is the standard deviation of the residuals, $\bar{w} \coloneqq \sum_{t'= t-w}^{t} t'/(w+1) = t - w/2$ is the window mean, and $s^2_w \coloneqq \sum_{t'= t-w}^{t} (t'-\bar{w})^2/w = (w+1)(w+2)/12$ is the window variance. To understand how the window size and prediction horizon affect the interval, consider larger windows, $w \geq 10$, and relatively long-horizon predictions with $d(t') \coloneqq (t'-t) \geq 2w$. In this regime, we can further approximate the interval as
\begin{align}
    \tilde{s}^i_t(t') \approx s^i_t \sqrt{1 + \frac{12 d^2(t')}{w}}.
\end{align}
This expression indicates that the size of the prediction interval decreases as the window size increases, since more data provides greater confidence in the estimates, and increases as the relative prediction horizon grows.

\subsection{Policy Selection}

Policy selection is performed through an iterative fine-tuning loop. At each iteration, we select a policy for fine-tuning according to the UCBs of future policy values forecasted by the fitted local linear models, fine-tune the most recent policy in the corresponding lineage, and evaluate the resulting policy through online interaction. This process is repeated until the interaction budget is exhausted, after which we return the policy with the highest obtained value estimate. The intuition behind this procedure is to efficiently allocate fine-tuning resources while maintaining a reasonable regret bound.

At iteration $j$, for each configuration index $i$, we store the value estimates obtained through online evaluation in a list $\mathbf{v}_j^i$ and fit a local linear model to predict future values using a fixed window size $w$. During the initial stage of fine-tuning, however, the window size is capped by the number of previous value estimates. This produces larger prediction intervals, thereby encouraging exploration especially during the initial phase. We predict the value $w$ iterations into the future whenever the remaining interaction budget permits; otherwise, we predict the value at the final iteration allowed by the remaining budget.

Let $t_j^i$ denote the number of policies in the lineage associated with the $i$th configuration at fine-tuning iteration $j$. We project the future value $\tilde{v}_{t'}^i$ and its associated prediction interval $\tilde{s}_{t'}^i$ using the locally fitted linear model described above, where $t' \coloneqq t + \min\{w, N-j\}$, and compute the UCB as
\begin{align}
    u^i_j \coloneqq \tilde{v}^i_{t'} + c \tilde{s}^i_{t'}
\end{align}
where $c$ is a parameter that scales the prediction interval and thereby controls the exploration-exploitation trade-off. We then select the policy with the largest UCB,
\begin{align}
    i^*_j \coloneqq \mathrm{argmax}_i u^i_j
\end{align}
fine-tune the most recent policy in the selected lineage, and append the estimated value of the resulting policy to the corresponding value list.

Inspired by the illustrative example in Fig.~\ref{fig:problem}, we propose a procedure for determining the exploration degree based on the highest policy value estimate obtained up to iteration $j$, defined as
\begin{align}
    v^*_j \coloneqq \max_i \max_{0 \leq t \leq t^i_j} \hat{v}^i_t,
\end{align}
Intuitively, our procedure increases the scale of the prediction intervals until one of the UCBs exceeds $v^*_j$ and picks its corresponding policy. In this way, we avoid introducing $c$ as an additional tunable parameter. Specifically, we select
\begin{align}
      i_j^* \coloneqq \mathrm{argmin}_i (v^*_j - \tilde{v}^i_{t'})/\tilde{s}^i_{t'}.
\end{align}

Overall, our active policy-selection strategy is based on two key principles: (i) short-horizon forecasting based on local trends to allocate the fine-tuning budget efficiently and (ii) UCB-based, automatically calibrated exploration to avoid prematurely committing to suboptimal policy lineages.

\section{Experiments}

\subsection{Implementation Details}

We employ four representative training algorithms designed for offline pretraining followed by online fine-tuning: AWAC \cite{nair2020awac}, IQL \cite{kostrikov2021offline}, CalQL \cite{nakamoto2023cal}, and ReBRAC \cite{tarasov2023revisiting}. We use the implementations of these algorithms provided by the offline RL library \textsc{d3rlpy} \cite{d3rlpy}. For each algorithm, we consider four hyperparameter settings obtained from the Cartesian product of two batch sizes (the default and half the default) and two learning rates (the default and half the default). We pretrain one candidate policy for each setting for 200 K iterations on the corresponding dataset, while fixing all remaining hyperparameters to the \textsc{d3rlpy} defaults. This procedure yields 16 pretrained candidate policies for each environment. We conduct the simulated experiments on a Linux machine using 16 AMD EPYC processor cores and 32 GB of RAM for each environment. We repeat each experiment using four random seeds.

We fine-tune the pretrained policies through online interactions with the environments, using the same hyperparameters as during pretraining. At each fine-tuning iteration, we use 4 K online transitions for training and an additional 1 K transitions for evaluation, resulting in a total cost of 5 K transitions per iteration. We consider window sizes of $w \in \{3,4,5,6,7\}$ for the value estimates used to fit the local linear model and generate forecasts. We use $w=3$ for the main experiments and evaluate the remaining window sizes in the ablation study.

\subsection{Baselines}

We compare our adaptive policy-selection and fine-tuning approach with several standard O2O-RL baselines. \textsc{random}: A pretrained policy is selected uniformly at random and used without fine-tuning. \textsc{best}: The pretrained policy with the highest estimated value is selected without fine-tuning. \textsc{fts} (Fine-Tuning Single): A randomly selected pretrained policy is fine-tuned using the entire interaction budget, after which the policy with the highest value in its lineage is selected. \textsc{fta} (Fine-Tuning All): All pretrained policies are fine-tuned by dividing the interaction budget equally among them, after which the policy with the highest estimated value is selected. \textsc{active} (Ours): Our active O2O-RL approach; see Algorithm~\ref{alg:active}.

\subsection{Environments}

We evaluate our approach on a diverse set of continuous-control environments from the Minari Python library \cite{farama2021}, spanning navigation, classic-control, and locomotion tasks. Minari provides datasets generated using MuJoCo \cite{todorov2012mujoco} in accordance with the principles of the D4RL benchmark \cite{fu2020d4rl}. Further details on dataset generation are provided in \cite{farama2021}. For each environment, we construct the downstream fine-tuning task by fixing the initial state using a predetermined random seed.
For the navigation tasks, we consider the Point Maze (\textsc{Maze}) environments with dense, distance-based rewards. The observations include both state and goal information, whereas the actions correspond to planar forces. We use datasets for open, U-shaped, and medium mazes, each containing 1,000 K transitions.

We additionally consider several classic-control environments. In InvertedDoublePendulum (\textsc{Pendulum}), a cart must balance a two-link pendulum, with rewards encouraging upright and stable behavior. In Swimmer (\textsc{Swimmer}), a three-link agent must move forward using joint torques, with rewards balancing forward progress against control costs. In Reacher (\textsc{Reacher}), a two-link robotic arm must move its fingertip toward a target, with rewards penalizing both the distance to the target and the control effort. In Pusher (\textsc{Pusher}), a robotic arm must push an object toward a goal, with rewards penalizing the object-goal distance, the fingertip-object distance, and the control effort. For each environment, we use datasets collected by medium and expert policies. The \textsc{Pendulum} datasets contain 100 K transitions; the \textsc{Reacher} and \textsc{Pusher} datasets each contain 500 K transitions; and the \textsc{Swimmer} datasets contain 1,000 K transitions.

Finally, we consider four standard legged-robot locomotion environments that are widely used to evaluate offline RL algorithms: Hopper (\textsc{Hopper}), HalfCheetah (\textsc{Cheetah}), Walker2d (\textsc{Walker}), and Ant (\textsc{Ant}). In these environments, a planar one-legged robot, a planar cheetah-like robot, a planar bipedal robot, and a quadrupedal robot, respectively, must move forward by applying joint torques. The observations describe the positions and velocities of the body parts, whereas the rewards combine forward velocity, a healthy-state bonus, and a control penalty. Each environment provides three datasets collected by simple, medium, and expert policies, with each dataset containing 1,000 K transitions.

\begin{table*}[htbp]
\begin{center}

\caption{Main Results.} \label{table:main} 
\begin{adjustbox}{width=0.81\textwidth}
\begin{tabular}{cc|cc|ccc}
\multicolumn{2}{c|}{\multirow{2}{*}{Environment}}  & \multicolumn{2}{c|}{Offline} & \multicolumn{3}{c}{Online} \\ \cline{3-7} 
& & \multicolumn{1}{c|}{\textsc{random}} & \multicolumn{1}{c|}{\textsc{best}} & \multicolumn{1}{c|}{\textsc{fts}} & \multicolumn{1}
{c|}{\textsc{fta}} & \textsc{active} \\ \hline\hline

\parbox[t]{2mm}{\multirow{3}{*}{\rotatebox[origin=c]{90}{\textsc{maze}}}}
& \textsc{open-dense} & -8.5 $\pm$  5.9 & 44.5 $\pm$  30.5 & 82.3 $\pm$  1.2 & 94.8 $\pm$  4.1 & \textbf{95.7}$\pm$  4.5 \\
& \textsc{umaze-dense} & -19.6 $\pm$  16.4 & 1.5 $\pm$  12.6 & 66.2 $\pm$  15.8 & 92.9 $\pm$  5.5 & \textbf{97.5}$\pm$  2.0 \\
& \textsc{medium-dense} & -11.9 $\pm$  15.8 & -1.4 $\pm$  28.4 & 52.5 $\pm$  6.7 & 96.7 $\pm$  2.1 & \textbf{98.6}$\pm$  0.9 \\
\hline
& \textit{Average} & -13.3 $\pm$  10.9 & 14.8 $\pm$  15.3 & 67.0 $\pm$  3.1 & 94.8 $\pm$  2.3 & \textbf{97.3}$\pm$  2.0 \\
\hline\hline

\parbox[t]{2mm}{\multirow{2}{*}{\rotatebox[origin=c]{90}{\textsc{pen.}}}}
& \textsc{medium} & 0.2 $\pm$  0.0 & 0.8 $\pm$  0.3 & 83.1 $\pm$  6.7 & \textbf{100.0}$\pm$  0.0 & \textbf{100.0}$\pm$  0.0 \\
& \textsc{expert} & 0.2 $\pm$  0.0 & 0.6 $\pm$  0.1 & 95.4 $\pm$  5.1 & \textbf{100.0}$\pm$  0.0 & \textbf{100.0}$\pm$  0.0 \\
\hline
& \textit{Average} & 0.2 $\pm$  0.0 & 0.7 $\pm$  0.2 & 89.3 $\pm$  5.1 & \textbf{100.0}$\pm$  0.0 & \textbf{100.0}$\pm$  0.0 \\
\hline\hline

\parbox[t]{2mm}{\multirow{2}{*}{\rotatebox[origin=c]{90}{\textsc{swi.}}}}
& \textsc{medium} & 22.6 $\pm$  2.2 & 36.9 $\pm$  3.4 & 46.6 $\pm$  3.2 & 48.8 $\pm$  6.9 & \textbf{61.4}$\pm$  17.7 \\
& \textsc{expert} & -11.6 $\pm$  6.0 & 54.3 $\pm$  27.1 & 51.2 $\pm$  7.6 & 54.0 $\pm$  6.6 & \textbf{67.8}$\pm$  20.2 \\
\hline
& \textit{Average} & 5.5 $\pm$  3.5 & 45.6 $\pm$  14.1 & 48.9 $\pm$  3.3 & 51.4 $\pm$  5.2 & \textbf{64.6}$\pm$  11.6 \\
\hline\hline

\parbox[t]{2mm}{\multirow{2}{*}{\rotatebox[origin=c]{90}{\textsc{rea.}}}}
& \textsc{medium} & 60.3 $\pm$  29.6 & 81.4 $\pm$  6.1 & 75.4 $\pm$  36.6 & 98.8 $\pm$  2.0 & \textbf{99.9}$\pm$  0.0 \\
& \textsc{expert} & 83.0 $\pm$  11.6 & 86.2 $\pm$  11.0 & 97.0 $\pm$  2.0 & \textbf{100.0}$\pm$  0.0 & \textbf{99.9}$\pm$  0.0 \\
\hline
& \textit{Average} & 71.6 $\pm$  11.3 & 83.8 $\pm$  7.6 & 86.2 $\pm$  17.4 & 99.4 $\pm$  1.0 & \textbf{99.9}$\pm$  0.0 \\
\hline\hline

\parbox[t]{2mm}{\multirow{2}{*}{\rotatebox[origin=c]{90}{\textsc{pus.}}}}
& \textsc{medium} & 80.7 $\pm$  0.9 & 85.2 $\pm$  1.7 & 91.3 $\pm$  0.4 & \textbf{95.0}$\pm$  1.1 & \textbf{95.0}$\pm$  0.6 \\
& \textsc{expert} & 74.7 $\pm$  1.5 & 84.6 $\pm$  1.8 & 92.8 $\pm$  1.4 & 93.2 $\pm$  1.7 & \textbf{94.2}$\pm$  1.2 \\
\hline
& \textit{Average} & 77.7 $\pm$  1.2 & 84.9 $\pm$  1.6 & 92.0 $\pm$  0.7 & 94.1 $\pm$  1.3 & \textbf{94.6}$\pm$  0.5 \\
\hline\hline

\parbox[t]{2mm}{\multirow{3}{*}{\rotatebox[origin=c]{90}{\textsc{hopp.}}}}
& \textsc{simple} & -0.3 $\pm$  0.0 & 0.5 $\pm$  0.5 & 44.9 $\pm$  4.7 & 36.1 $\pm$  9.4 & \textbf{96.5}$\pm$  2.1 \\
& \textsc{medium} & -0.0 $\pm$  0.2 & 2.6 $\pm$  2.6 & 58.9 $\pm$  4.9 & 29.7 $\pm$  0.6 & \textbf{72.9}$\pm$  25.0 \\
& \textsc{expert} & -0.2 $\pm$  0.0 & 0.1 $\pm$  0.1 & 55.2 $\pm$  1.6 & 35.5 $\pm$  10.2 & \textbf{92.3}$\pm$  3.3 \\
\hline
& \textit{Average} & -0.2 $\pm$  0.1 & 1.1 $\pm$  0.8 & 53.0 $\pm$  1.7 & 33.8 $\pm$  6.4 & \textbf{87.2}$\pm$  9.0 \\
\hline\hline

\parbox[t]{2mm}{\multirow{3}{*}{\rotatebox[origin=c]{90}{\textsc{chee.}}}}
& \textsc{simple} & -0.2 $\pm$  0.5 & 1.6 $\pm$  0.8 & 46.2 $\pm$  2.8 & 44.2 $\pm$  3.4 & \textbf{83.8}$\pm$  9.6 \\
& \textsc{medium} & 0.0 $\pm$  0.1 & 2.1 $\pm$  0.4 & 49.7 $\pm$  2.8 & 39.6 $\pm$  4.5 & \textbf{53.9}$\pm$  12.8 \\
& \textsc{expert} & 1.3 $\pm$  0.2 & 2.8 $\pm$  0.4 & 54.8 $\pm$  3.0 & 44.3 $\pm$  2.2 & \textbf{87.1}$\pm$  9.4 \\
\hline
& \textit{Average} & 0.4 $\pm$  0.2 & 2.2 $\pm$  0.1 & 50.2 $\pm$  1.4 & 42.7 $\pm$  1.7 & \textbf{74.9}$\pm$  10.2 \\
\hline\hline

\parbox[t]{2mm}{\multirow{3}{*}{\rotatebox[origin=c]{90}{\textsc{walk.}}}}
& \textsc{simple} & 2.6 $\pm$  0.2 & 7.5 $\pm$  0.5 & 46.9 $\pm$  2.4 & 35.1 $\pm$  8.1 & \textbf{70.9}$\pm$  28.2 \\
& \textsc{medium} & 2.5 $\pm$  0.4 & 8.7 $\pm$  3.4 & 53.7 $\pm$  5.7 & 22.9 $\pm$  2.8 & \textbf{74.3}$\pm$  12.1 \\
& \textsc{expert} & 5.0 $\pm$  0.7 & 7.9 $\pm$  0.8 & 54.2 $\pm$  2.3 & 31.0 $\pm$  6.3 & \textbf{60.9}$\pm$  21.0 \\
\hline
& \textit{Average} & 3.4 $\pm$  0.3 & 8.0 $\pm$  1.4 & 51.6 $\pm$  2.2 & 29.6 $\pm$  1.4 & \textbf{68.7}$\pm$  12.2 \\
\hline\hline

\parbox[t]{2mm}{\multirow{3}{*}{\rotatebox[origin=c]{90}{\textsc{ant}}}}
& \textsc{simple} & -0.3 $\pm$  3.6 & 11.5 $\pm$  0.3 & 38.8 $\pm$  1.2 & 27.0 $\pm$  4.8 & \textbf{67.1}$\pm$  18.1 \\
& \textsc{medium} & -1.4 $\pm$  1.9 & 16.6 $\pm$  0.7 & 35.2 $\pm$  2.5 & 16.5 $\pm$  1.2 & \textbf{42.9}$\pm$  25.0 \\
& \textsc{expert} & 6.6 $\pm$  0.9 & 19.3 $\pm$  4.1 & 46.6 $\pm$  8.9 & 23.1 $\pm$  7.3 & \textbf{50.9}$\pm$  11.1 \\
\hline
& \textit{Average} & 1.6 $\pm$  0.8 & 15.8 $\pm$  1.1 & 40.2 $\pm$  2.6 & 22.2 $\pm$  2.7 & \textbf{53.6}$\pm$  10.9 \\
\hline\hline

& \textit{Overall Average} & 16.3 $\pm$  2.1 & 28.6 $\pm$  2.4 & 64.3 $\pm$  2.0 & 63.1 $\pm$  0.7 & \textbf{82.3}$\pm$  1.9

\end{tabular}
\end{adjustbox}

\end{center}
\end{table*}

\begin{table*}[htbp]
\begin{center}

\caption{Results for Smaller Budget.} \label{table:small} 
\begin{adjustbox}{width=0.81\textwidth}
\begin{tabular}{cc|cc|ccc}
\multicolumn{2}{c|}{\multirow{2}{*}{Environment}}  & \multicolumn{2}{c|}{Offline} & \multicolumn{3}{c}{Online} \\ \cline{3-7} 
& & \multicolumn{1}{c|}{\textsc{random}} & \multicolumn{1}{c|}{\textsc{best}} & \multicolumn{1}{c|}{\textsc{fts}} & \multicolumn{1}
{c|}{\textsc{fta}} & \textsc{active} \\ \hline\hline

\parbox[t]{2mm}{\multirow{3}{*}{\rotatebox[origin=c]{90}{\textsc{maze}}}}
& \textsc{open-dense} & -8.5 $\pm$  5.9 & 44.5 $\pm$  30.5 & 81.9 $\pm$  1.3 & 94.1 $\pm$  4.2 & \textbf{95.3}$\pm$  4.4 \\
& \textsc{umaze-dense} & -19.6 $\pm$  16.4 & 1.5 $\pm$  12.7 & 66.0 $\pm$  15.7 & 92.4 $\pm$  5.2 & \textbf{97.4}$\pm$  2.0 \\
& \textsc{medium-dense} & -11.9 $\pm$  15.8 & -1.4 $\pm$  28.4 & 51.7 $\pm$  6.7 & 96.3 $\pm$  1.7 & \textbf{98.4}$\pm$  0.7 \\
\hline
& \textit{Average} & -13.4 $\pm$  10.9 & 14.9 $\pm$  15.3 & 66.5 $\pm$  3.1 & 94.3 $\pm$  2.4 & \textbf{97.0}$\pm$  2.1 \\
\hline\hline

\parbox[t]{2mm}{\multirow{2}{*}{\rotatebox[origin=c]{90}{\textsc{pen.}}}}
& \textsc{medium} & 0.2 $\pm$  0.0 & 0.8 $\pm$  0.3 & 83.1 $\pm$  6.7 & \textbf{100.0}$\pm$  0.0 & \textbf{100.0}$\pm$  0.0 \\
& \textsc{expert} & 0.2 $\pm$  0.0 & 0.6 $\pm$  0.1 & 91.3 $\pm$  9.6 & 80.9 $\pm$  33.1 & \textbf{100.0}$\pm$  0.0 \\
\hline
& \textit{Average} & 0.2 $\pm$  0.0 & 0.7 $\pm$  0.2 & 87.2 $\pm$  7.1 & 90.4 $\pm$  16.5 & \textbf{100.0}$\pm$  0.0 \\
\hline\hline

\parbox[t]{2mm}{\multirow{2}{*}{\rotatebox[origin=c]{90}{\textsc{swi.}}}}
& \textsc{medium} & 23.0 $\pm$  2.4 & 37.7 $\pm$  4.0 & 45.1 $\pm$  4.0 & 49.5 $\pm$  7.3 & \textbf{51.1}$\pm$  6.6 \\
& \textsc{expert} & -11.7 $\pm$  5.9 & 55.0 $\pm$  26.5 & 50.2 $\pm$  6.6 & 55.3 $\pm$  5.8 & \textbf{56.1}$\pm$  6.7 \\
\hline
& \textit{Average} & 5.7 $\pm$  3.6 & 46.4 $\pm$  13.7 & 47.7 $\pm$  1.7 & 52.4 $\pm$  5.1 & \textbf{53.6}$\pm$  4.5 \\
\hline\hline

\parbox[t]{2mm}{\multirow{2}{*}{\rotatebox[origin=c]{90}{\textsc{rea.}}}}
& \textsc{medium} & 60.3 $\pm$  29.6 & 81.4 $\pm$  6.1 & 75.3 $\pm$  36.5 & 98.8 $\pm$  2.0 & \textbf{99.9}$\pm$  0.0 \\
& \textsc{expert} & 83.0 $\pm$  11.6 & 86.2 $\pm$  11.0 & 97.0 $\pm$  2.0 & \textbf{100.0}$\pm$  0.0 & \textbf{99.9}$\pm$  0.1 \\
\hline
& \textit{Average} & 71.7 $\pm$  11.3 & 83.8 $\pm$  7.6 & 86.2 $\pm$  17.4 & 99.4 $\pm$  1.0 & \textbf{99.9}$\pm$  0.1 \\
\hline\hline

\parbox[t]{2mm}{\multirow{2}{*}{\rotatebox[origin=c]{90}{\textsc{pus.}}}}
& \textsc{medium} & 81.1 $\pm$  1.2 & 85.6 $\pm$  1.7 & 91.2 $\pm$  0.5 & \textbf{94.8}$\pm$  1.0 & \textbf{94.8}$\pm$  1.0 \\
& \textsc{expert} & 75.2 $\pm$  1.6 & 85.1 $\pm$  1.6 & 92.9 $\pm$  1.2 & 93.6 $\pm$  1.5 & \textbf{94.3}$\pm$  1.0 \\
\hline
& \textit{Average} & 78.1 $\pm$  1.4 & 85.3 $\pm$  1.5 & 92.0 $\pm$  0.6 & 94.2 $\pm$  1.1 & \textbf{94.5}$\pm$  0.6 \\
\hline\hline

\parbox[t]{2mm}{\multirow{3}{*}{\rotatebox[origin=c]{90}{\textsc{hopp.}}}}
& \textsc{simple} & -0.3 $\pm$  0.0 & 0.5 $\pm$  0.5 & 41.1 $\pm$  4.9 & 31.1 $\pm$  0.6 & \textbf{82.9}$\pm$  24.4 \\
& \textsc{medium} & -0.0 $\pm$  0.2 & 2.7 $\pm$  2.7 & \textbf{55.4}$\pm$  3.6 & 30.1 $\pm$  0.6 & 45.7 $\pm$  22.6 \\
& \textsc{expert} & -0.2 $\pm$  0.0 & 0.1 $\pm$  0.1 & 53.1 $\pm$  4.3 & 35.8 $\pm$  10.2 & \textbf{78.1}$\pm$  12.1 \\
\hline
& \textit{Average} & -0.2 $\pm$  0.1 & 1.1 $\pm$  0.8 & 49.9 $\pm$  0.7 & 32.3 $\pm$  3.1 & \textbf{68.9}$\pm$  12.0 \\
\hline\hline

\parbox[t]{2mm}{\multirow{3}{*}{\rotatebox[origin=c]{90}{\textsc{chee.}}}}
& \textsc{simple} & -0.2 $\pm$  0.6 & 1.7 $\pm$  0.8 & 44.7 $\pm$  2.9 & 39.7 $\pm$  2.7 & \textbf{80.2}$\pm$  12.4 \\
& \textsc{medium} & 0.0 $\pm$  0.1 & 2.3 $\pm$  0.5 & \textbf{48.7}$\pm$  4.3 & 35.1 $\pm$  10.8 & 41.1 $\pm$  14.5 \\
& \textsc{expert} & 1.4 $\pm$  0.2 & 3.0 $\pm$  0.4 & 53.8 $\pm$  2.9 & 41.6 $\pm$  2.6 & \textbf{87.4}$\pm$  9.2 \\
\hline
& \textit{Average} & 0.4 $\pm$  0.2 & 2.3 $\pm$  0.1 & 49.1 $\pm$  2.2 & 38.8 $\pm$  3.6 & \textbf{69.5}$\pm$  9.0 \\
\hline\hline

\parbox[t]{2mm}{\multirow{3}{*}{\rotatebox[origin=c]{90}{\textsc{walk.}}}}
& \textsc{simple} & 2.7 $\pm$  0.3 & 7.8 $\pm$  0.7 & 46.0 $\pm$  1.4 & 24.1 $\pm$  1.4 & \textbf{67.0}$\pm$  25.8 \\
& \textsc{medium} & 2.6 $\pm$  0.4 & 9.0 $\pm$  3.5 & 51.8 $\pm$  3.9 & 23.3 $\pm$  3.0 & \textbf{59.4}$\pm$  19.4 \\
& \textsc{expert} & 5.3 $\pm$  0.6 & 8.3 $\pm$  0.8 & 54.0 $\pm$  1.4 & 27.0 $\pm$  4.1 & \textbf{55.5}$\pm$  20.7 \\
\hline
& \textit{Average} & 3.6 $\pm$  0.3 & 8.4 $\pm$  1.5 & 50.6 $\pm$  1.4 & 24.8 $\pm$  1.5 & \textbf{60.6}$\pm$  17.0 \\
\hline\hline

\parbox[t]{2mm}{\multirow{3}{*}{\rotatebox[origin=c]{90}{\textsc{ant}}}}
& \textsc{simple} & -0.3 $\pm$  3.9 & 12.5 $\pm$  0.6 & 37.2 $\pm$  0.5 & 24.0 $\pm$  3.7 & \textbf{61.1}$\pm$  19.3 \\
& \textsc{medium} & -1.6 $\pm$  2.1 & 18.5 $\pm$  0.7 & \textbf{35.5}$\pm$  1.9 & 18.1 $\pm$  1.1 & \textbf{35.5}$\pm$  20.2 \\
& \textsc{expert} & 7.2 $\pm$  1.2 & 20.8 $\pm$  4.1 & \textbf{46.0}$\pm$  8.8 & 20.8 $\pm$  4.2 & 37.9 $\pm$  13.5 \\
\hline
& \textit{Average} & 1.8 $\pm$  0.9 & 17.3 $\pm$  1.0 & 39.6 $\pm$  2.5 & 21.0 $\pm$  2.0 & \textbf{44.8}$\pm$  5.5 \\
\hline\hline

& \textit{Overall Average} & 16.4 $\pm$  2.1 & 28.9 $\pm$  2.4 & 63.2 $\pm$  1.7 & 60.8 $\pm$  1.7 & \textbf{76.6}$\pm$  1.5

\end{tabular}
\end{adjustbox}

\end{center}
\end{table*}

\subsection{Main Results}

We conduct experiments using an online interaction budget of 1,000 K transitions for each dataset. The main results are presented in Table~\ref{table:main}. We compare the O2O-RL approaches based on the value of the final fine-tuned policy selected by each approach. For each environment, we rescale the mean returns using min-max normalization. The minimum is set to the estimated value of a random policy evaluated in the environment, while the maximum is set to the value of the best-performing policy ($v^{i^*}_{j^*}$; see Problem~\ref{problem}) that could be obtained if future fine-tuning values were known \emph{a priori}. We repeat each experiment using four random seeds and report the mean values and corresponding standard deviations as percentages.

Across the environments and datasets, offline RL approaches typically achieve lower scores and therefore incur greater opportunity loss, or regret, relative to the maximum achievable score of 100\% obtained through fine-tuning. These results underscore the importance of effectively using the online interaction budget. For instance, in the \textsc{maze} environments, the pretrained policies perform worse on average (\textsc{random}) than a policy that takes random actions. Even the best pretrained policy (\textsc{best}) can perform worse than a random policy, as observed in the \textsc{maze-medium-dense} environment. This finding does not necessarily imply that offline training is ineffective. A small number of fine-tuning iterations can substantially improve the performance of pretrained policies, often more rapidly than training from scratch. Without fine-tuning, however, this potential may remain unrealized.

The baseline fine-tuning approaches, \textsc{fts} and \textsc{fta}, substantially outperform the offline approaches. Committing to a single policy, as in \textsc{fts}, often yields considerable performance gains because the entire interaction budget is devoted to fine-tuning that policy, as observed in the legged-robot locomotion environments. However, if the selected policy becomes trapped in a local optimum or improves slowly, the entire budget may be spent on a suboptimal policy, as observed in the \textsc{swimmer} environment. By contrast, \textsc{fta} performs better when the interaction budget is relatively large for a given environment, as observed in \textsc{pendulum}, \textsc{reacher}, and \textsc{pusher}. However, \textsc{fta} struggles in environments such as the legged-robot locomotion tasks, where substantial performance gains emerge only after a relatively long fine-tuning warm-up period.

Our approach, \textsc{active}, actively selects and fine-tunes policies, effectively combining the strengths of both baselines. For example, in the \textsc{maze} environments, \textsc{active} does not remain committed to policy lineages that fail to improve. Instead, it switches to and fine-tunes other policies, exploring a sufficient portion of the candidate pool and thereby identifying a promising policy lineage, unlike \textsc{fts}. It then uses the remaining budget for further fine-tuning, ultimately achieving higher scores than \textsc{fta}. In the legged-robot locomotion environments, \textsc{active} avoids spending the entire budget on fine-tuning every candidate, particularly when the candidates are weakly pretrained and require long warm-up periods. Instead, it successfully identifies promising policy lineages that improve during fine-tuning and allocates most of the interaction budget to them. This allocation allows it to overcome the long warm-up period, unlike \textsc{fta}. Moreover, when progress slows, \textsc{active} switches among promising policies, enabling it to achieve higher scores than \textsc{fts}.

Nevertheless, \textsc{active} struggles in environments such as \textsc{swimmer} and \textsc{ant}, highlighting an important limitation. In these environments, the policy lineages show little or no indication of improvement for extended periods during the initial phase of fine-tuning, as illustrated for Seed~3 in Fig.~\ref{fig:finetuning}. Consequently, our approach may exhaust the entire interaction budget while attempting to fine-tune all candidate policies in order to identify one that improves. This limitation could potentially be mitigated by incorporating a consistent risk-taking objective when none of the policy lineages are improving, while preserving the exploration needed to distinguish among candidate policies. However, incorporating such an objective could make the overall approach brittle and sensitive to algorithmic choices. We therefore leave this direction for future work.

\begin{table*}[htbp]
\begin{center}

\caption{Ablation Study of  Window Size} \label{table:ablation} 
\begin{adjustbox}{width=0.75\textwidth}
\begin{tabular}{cc|cccc}
\multicolumn{2}{c|}{\multirow{2}{*}{Environment}}  & \multicolumn{4}{c}{Locality Window Size ($w$)} \\ \cline{3-6} 
& & $w=3$ & $w=5$ & $w=6$ & $w=7$ 
\\ \hline\hline

\parbox[t]{2mm}{\multirow{3}{*}{\rotatebox[origin=c]{90}{\textsc{maze}}}}
& \textsc{open-dense} & 0.6 $\pm$  0.6 & -1.1 $\pm$  3.5 & 0.8 $\pm$  1.0 & 0.7 $\pm$  1.1 \\
& \textsc{umaze-dense} & 0.0 $\pm$  0.8 & -0.0 $\pm$  0.4 & 0.2 $\pm$  0.6 & 0.3 $\pm$  0.5 \\
& \textsc{medium-dense} & -0.1 $\pm$  0.6 & -0.1 $\pm$  0.4 & 0.1 $\pm$  0.7 & 0.5 $\pm$  0.5 \\
\hline
& \textit{Average} & 0.2 $\pm$  0.6 & -0.4 $\pm$  1.2 & 0.4 $\pm$  0.6 & 0.5 $\pm$  0.5 \\
\hline\hline

\parbox[t]{2mm}{\multirow{2}{*}{\rotatebox[origin=c]{90}{\textsc{pen.}}}}
& \textsc{medium} & -0.0 $\pm$  0.0 & 0.0 $\pm$  0.0 & 0.0 $\pm$  0.0 & 0.0 $\pm$  0.0 \\
& \textsc{expert} & 0.0 $\pm$  0.0 & 0.0 $\pm$  0.0 & 0.0 $\pm$  0.0 & 0.0 $\pm$  0.0 \\
\hline
& \textit{Average} & -0.0 $\pm$  0.0 & 0.0 $\pm$  0.0 & 0.0 $\pm$  0.0 & 0.0 $\pm$  0.0 \\
\hline\hline

\parbox[t]{2mm}{\multirow{2}{*}{\rotatebox[origin=c]{90}{\textsc{swi.}}}}
& \textsc{medium} & -9.8 $\pm$  14.4 & -0.8 $\pm$  1.4 & -0.3 $\pm$  0.5 & -4.7 $\pm$  6.1 \\
& \textsc{expert} & -12.0 $\pm$  15.7 & -0.8 $\pm$  3.2 & -2.2 $\pm$  5.1 & -2.9 $\pm$  26.0 \\
\hline
& \textit{Average} & -10.9 $\pm$  8.4 & -0.8 $\pm$  1.7 & -1.2 $\pm$  2.6 & -3.8 $\pm$  13.0 \\
\hline\hline

\parbox[t]{2mm}{\multirow{2}{*}{\rotatebox[origin=c]{90}{\textsc{rea.}}}}
& \textsc{medium} & 0.0 $\pm$  0.0 & 0.0 $\pm$  0.0 & 0.0 $\pm$  0.0 & 0.0 $\pm$  0.0 \\
& \textsc{expert} & -0.0 $\pm$  0.0 & 0.0 $\pm$  0.0 & 0.0 $\pm$  0.0 & 0.0 $\pm$  0.0 \\
\hline
& \textit{Average} & -0.0 $\pm$  0.0 & 0.0 $\pm$  0.0 & 0.0 $\pm$  0.0 & 0.0 $\pm$  0.0 \\
\hline\hline

\parbox[t]{2mm}{\multirow{2}{*}{\rotatebox[origin=c]{90}{\textsc{pus.}}}}
& \textsc{medium} & -0.4 $\pm$  1.0 & 0.0 $\pm$  0.1 & 0.1 $\pm$  0.2 & 0.1 $\pm$  0.8 \\
& \textsc{expert} & -0.5 $\pm$  0.8 & -0.3 $\pm$  0.5 & -0.3 $\pm$  0.6 & -0.3 $\pm$  0.6 \\
\hline
& \textit{Average} & -0.4 $\pm$  0.8 & -0.1 $\pm$  0.2 & -0.1 $\pm$  0.2 & -0.1 $\pm$  0.5 \\
\hline\hline

\parbox[t]{2mm}{\multirow{3}{*}{\rotatebox[origin=c]{90}{\textsc{hopp.}}}}
& \textsc{simple} & -22.3 $\pm$  26.9 & -3.8 $\pm$  5.4 & -6.6 $\pm$  10.2 & -6.6 $\pm$  10.2 \\
& \textsc{medium} & -26.0 $\pm$  31.6 & -8.3 $\pm$  28.3 & -6.5 $\pm$  7.6 & -5.8 $\pm$  8.2 \\
& \textsc{expert} & -1.1 $\pm$  3.4 & 1.6 $\pm$  2.6 & -16.2 $\pm$  25.1 & -16.2 $\pm$  25.1 \\
\hline
& \textit{Average} & -16.4 $\pm$  17.7 & -3.5 $\pm$  8.3 & -9.8 $\pm$  6.5 & -9.5 $\pm$  7.7 \\
\hline\hline

\parbox[t]{2mm}{\multirow{3}{*}{\rotatebox[origin=c]{90}{\textsc{chee.}}}}
& \textsc{simple} & 0.8 $\pm$  4.1 & -0.9 $\pm$  1.5 & 4.7 $\pm$  10.3 & 4.2 $\pm$  9.5 \\
& \textsc{medium} & 6.2 $\pm$  11.8 & -11.0 $\pm$  15.0 & -17.9 $\pm$  17.6 & -17.9 $\pm$  17.6 \\
& \textsc{expert} & -2.6 $\pm$  2.7 & 0.0 $\pm$  0.0 & 0.4 $\pm$  0.7 & 7.2 $\pm$  11.5 \\
\hline
& \textit{Average} & 1.5 $\pm$  3.6 & -4.0 $\pm$  4.9 & -4.3 $\pm$  8.3 & -2.2 $\pm$  11.5 \\
\hline\hline

\parbox[t]{2mm}{\multirow{3}{*}{\rotatebox[origin=c]{90}{\textsc{walk.}}}}
& \textsc{simple} & -5.8 $\pm$  6.2 & -3.1 $\pm$  5.4 & 4.0 $\pm$  9.5 & 5.3 $\pm$  17.3 \\
& \textsc{medium} & -18.1 $\pm$  13.5 & 2.0 $\pm$  9.4 & 4.1 $\pm$  7.4 & 6.2 $\pm$  7.7 \\
& \textsc{expert} & 3.2 $\pm$  15.3 & 10.4 $\pm$  16.4 & 10.7 $\pm$  13.3 & 10.0 $\pm$  12.3 \\
\hline
& \textit{Average} & -6.9 $\pm$  9.0 & 3.1 $\pm$  9.2 & 6.3 $\pm$  7.7 & 7.2 $\pm$  9.9 \\
\hline\hline

\parbox[t]{2mm}{\multirow{3}{*}{\rotatebox[origin=c]{90}{\textsc{ant}}}}
& \textsc{simple} & -21.7 $\pm$  26.0 & -4.6 $\pm$  4.1 & -4.2 $\pm$  3.6 & -0.1 $\pm$  5.3 \\
& \textsc{medium} & -22.7 $\pm$  28.4 & -1.5 $\pm$  2.8 & -0.2 $\pm$  5.9 & 2.1 $\pm$  23.3 \\
& \textsc{expert} & -6.6 $\pm$  18.8 & -11.0 $\pm$  11.8 & -20.1 $\pm$  12.5 & -20.0 $\pm$  12.3 \\
\hline
& \textit{Average} & -17.0 $\pm$  16.3 & -5.7 $\pm$  4.1 & -8.2 $\pm$  5.4 & -6.0 $\pm$  10.1 \\
\hline\hline

& \textit{Overall Average} & -5.6 $\pm$  1.9 & -1.3 $\pm$  1.4 & -1.9 $\pm$  2.6 & -1.5 $\pm$  3.5 

\end{tabular}
\end{adjustbox}

\end{center}
\end{table*}

\subsection{Smaller Budget}

We also conduct experiments using a smaller interaction budget of 750 K transitions, with the results presented in Table~\ref{table:small}. The offline stage is performed in the same manner as in the main experiments; variations in the normalized offline scores arise from changes in the maximum policy value achievable under the reduced budget. Overall, the offline scores are comparable to those reported in Table~\ref{table:main}.

As expected, the fine-tuning approaches achieve lower scores under the reduced interaction budget. The smaller budget prevents \textsc{fts} from fine-tuning its selected policy for enough iterations to attain higher values. It also forces \textsc{fta} to allocate fewer fine-tuning iterations to each candidate policy. For \textsc{active}, the reduced budget further exacerbates the limitation discussed above by decreasing the number of interactions available to identify an improving policy lineage. This effect is particularly pronounced in the legged-robot locomotion environments, where policies require long fine-tuning warm-up periods.

\subsection{Ablation Study of Window Size}

We perform an ablation study to investigate the sensitivity of our approach to the window-size meta-parameter. Specifically, we repeat the experiments using larger and smaller values of the window size, $w$, used to fit the local linear model and generate forecasts, while holding all other parameters fixed. The results are presented in Table~\ref{table:ablation}. Overall, our approach exhibits some sensitivity to the choice of $w$.

Increasing the window size has only a limited effect on performance in the low-dimensional environments, with the exception of \textsc{swimmer}. In contrast, the high-dimensional locomotion tasks are more sensitive to the choice of $w$, particularly \textsc{hopper} and \textsc{ant}.
The setting $w=3$ is the smallest window size for which a prediction interval can be constructed. Because the variance of the estimated linear-regression parameters can be relatively high for such a small window, performance decreases, particularly in environments where the value estimates fluctuate substantially during fine-tuning, as shown in Fig.~\ref{fig:finetuning}. Performance also generally decreases for larger window sizes. However, it improves for \textsc{ant}, where larger windows produce less noisy forecasts that more reliably identify improving policy lineages.

\section{Conclusion}

To our knowledge, we propose the first active O2O-RL framework that jointly performs policy selection and fine-tuning under a limited online interaction budget. We begin by training a diverse set of candidate policies with offline RL using multiple algorithms and hyperparameter configurations. Because pretrained policies may perform arbitrarily poorly and their performance during fine-tuning may stall or regress depending on the algorithm, hyperparameters, environment, and even random seed, we adaptively allocate scarce online interactions using a UCB criterion based on predicted future policy values. After each fine-tuning iteration, we evaluate the selected policy and update a local linear regression model to refine the corresponding value forecasts and confidence bounds. The framework switches policies whenever another policy attains a higher UCB, thereby allocating the interaction budget efficiently. Across multiple benchmarks, our adaptive policy-selection and fine-tuning procedure consistently outperforms existing O2O-RL baselines.

Despite outperforming strong baselines, our approach leaves substantial room for improvement. One promising direction is to leverage exploration rollouts to monitor fine-tuning progress with little or no additional online evaluation. Another direction is to incorporate similarities among policies, as suggested by \cite{konyushova2021active}. An orthogonal avenue is to develop OPE methods tailored specifically to O2O-RL, with greater emphasis on eliminating poor pretrained policies and identifying policies with strong fine-tuning potential.
We believe that our framework advances offline RL research by taking another step toward practical, deployable methods for real-world systems in which online interactions are costly or risky.

\begin{acks}
This work was supported by the Commonwealth Cyber Initiative HV-2Q25-035, HC-2Q25-033, and the Central Virginia Node under the award VV-1Q26-001.
\end{acks}

\printbibliography

\appendix

\end{document}